\documentclass[runningheads, orivec]{llncs}
\usepackage{amsmath}
\usepackage{tabularx}
\usepackage[T1]{fontenc}
\usepackage{float} 
\usepackage{graphicx,verbatim}
\usepackage{subcaption} 
\usepackage{caption}
\begin{document}

\title{Clinical Validation of Deep Learning for Real-Time Tissue Oxygenation Estimation Using Spectral Imaging}

\author{Jens De Winne\inst{1,2} \and
Siri Willems\inst{1} \and
Siri Luthman\inst{1} \and
Danilo Babin\inst{2} \and
Hiep Luong\inst{2}\ \and
Wim Ceelen\inst{3,4}}

\authorrunning{J. De Winne et al.}
\institute{Imec, Kapeldreef 75, 3001 Leuven, Belgium \and
Department of Telecommunications and Information Processing, Ghent University/imec, Sint-Pietersnieuwestraat 41, 9000 Ghent, Belgium \and
Department of Gastrointestinal Surgery, Ghent University Hospital, 9000 Ghent, Belgium \and
Department of Human Structure and Repair - Experimental Surgery Lab, Ghent University, 9000 Ghent, Belgium}
    
\maketitle{}

\begin{abstract}
Accurate, real-time monitoring of tissue ischemia is crucial to understand tissue health and guide surgery. Spectral imaging shows great potential for contactless and intraoperative monitoring of tissue oxygenation. Due to the difficulty of obtaining direct reference oxygenation values, conventional methods are based on linear unmixing techniques. These are prone to assumptions and these linear relations may not always hold in practice. In this work, we present deep learning approaches for real-time tissue oxygenation estimation using Monte-Carlo simulated spectra. We train a fully connected neural network (FCN) and a convolutional neural network (CNN) for this task and propose a domain-adversarial training approach to bridge the gap between simulated and real clinical spectral data. Results demonstrate that these deep learning models achieve a higher correlation with capillary lactate measurements, a well-known marker of hypoxia, obtained during spectral imaging in surgery, compared to traditional linear unmixing. Notably, domain-adversarial training effectively reduces the domain gap, optimizing performance in real clinical settings.

\keywords{multispectral imaging  \and tissue oxygenation \and domain adaptation \and deep learning \and simulations \and real-time imaging}
\end{abstract}

\section{Introduction}
The supply of oxygen-rich blood to tissue is essential for maintaining cellular homeostasis and metabolic processes, ensuring tissue health. Real-time monitoring is particularly crucial in intraoperative settings where rapid assessment of tissue oxygenation can guide surgical decisions and improve patient outcomes. Spectral imaging has emerged as a promising contactless technique for intraoperative tissue assessment in various surgical scenarios \cite{C1}. This advanced imaging technique captures data across multiple wavelengths of the electromagnetic spectrum, to create images representing the spectral properties of tissue in a spatial scene, and is called a three-dimensional hypercube (x,y,$\lambda$).\par

Many previous studies estimate tissue oxygenation from spectral data using linear unmixing. This technique decomposes each pixel's spectrum into a combination of individual tissue component spectra, assuming a linear relationship between them. However, in reality, the pixel spectra are influenced by environmental conditions and sensor limitations, and non linear relationships are expected. 

Advances in deep learning can cope with these non-linear relationships. A key challenge in training such models is the lack of ground truth oxygenation values in clinical practice. Therefore, Ayala et al. performed Monte-Carlo simulations to create labeled training data for the development of machine learning algorithms \cite{C2}. Modeling light-tissue interactions allows generating realistic spectral data under controlled conditions where direct oxygenation measurements are impractical. Translation of the developed machine learning models to clinical settings has already shown promising results in arm occlusion studies \cite{C3,C4} and partial nephrectomy \cite{C2}. Although the design of the simulations is carefully chosen, a mismatch between simulated and real data remains due to differences in optical properties, measurement noise, and imaging artifacts introduced by the spectral cameras. Dreher et al. proposed a conditional invertible neural network to close the domain gap and generate more realistic data from simulated labeled data, and evaluated on the downstream task of organ classification in pigs \cite{dreher2023}.\par

For the task of real-time oxygenation quantification in clinical practice, the main challenge remains to clinically validate the quantification methods. A common approach is to perform occlusion studies and qualitatively evaluate the time dynamics of the predictions. To mitigate this, we build on top of previously published literature to evaluate different deep learning approaches for real-time tissue oxygenation quantification in a clinical setting. We propose a clinical validation study using a real-time multispectral snapshot camera (MSI), comparing different methods including linear unmixing, neural networks, and domain adversarial (DA) networks. The predicted oxygenation values are evaluated by correlating them with corresponding capillary lactate measurements, a well-known marker of hypoxia.

\section{Data}

\subsection{Simulated data}
\label{sec:simulated-data}
\begin{figure}
    \centering
    \includegraphics[width=1\linewidth]{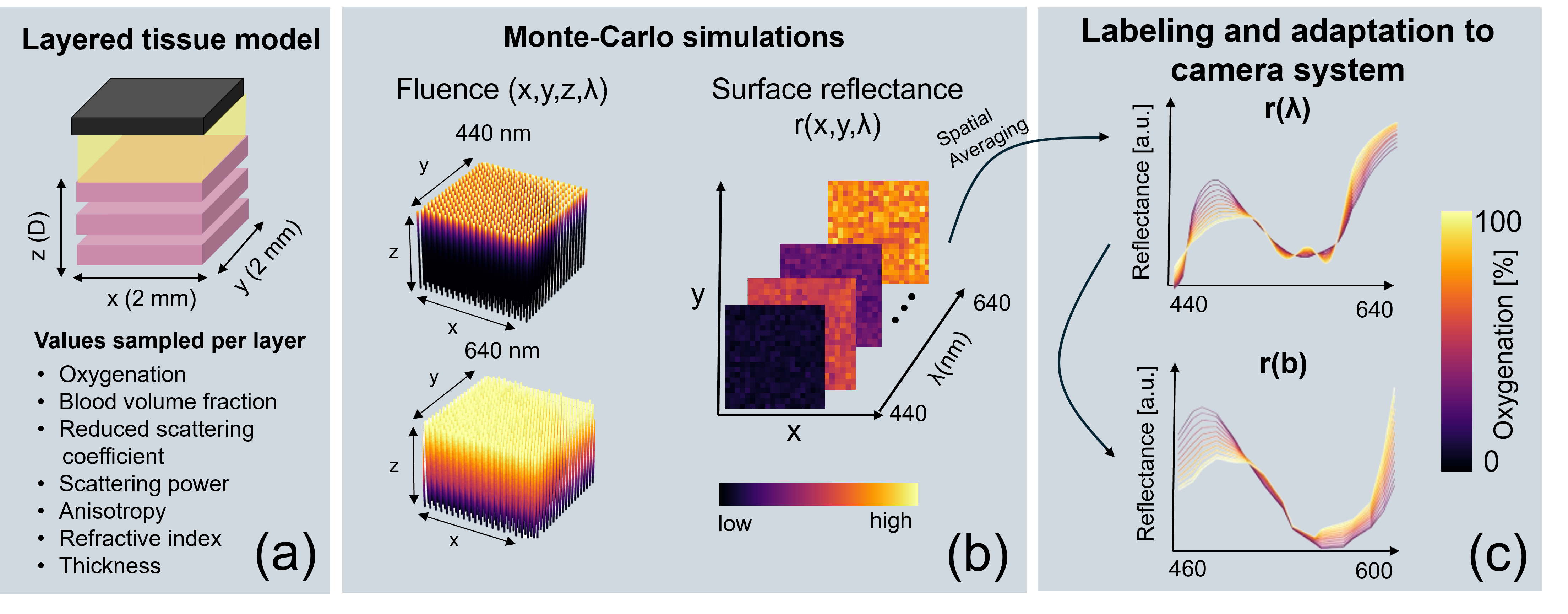}
    \caption{Overview of the approach to simulate oxygenation labeled reflectance spectra.}
    \label{fig:framework}
\end{figure}

Diffuse reflectance spectra are simulated in silico using Monte-Carlo simulations of light propagation in an artificial tissue. As tissues's oxygenation is defined within the simulation, reflectance spectra with a ground-truth oxygenation label are obtained and used to train the deep learning models. This simulation framework follows a similar approach as Ayala et al. \cite{C2,C11}. An overview is illustrated in Fig. \ref{fig:framework}.

Soft tissue typically consists of three distinct layers, from outermost to innermost: the serosa, muscularis and submucosa. Hence, we implement a three-dimensional 3-layered tissue model, see Fig. \ref{fig:framework}a. For each of the layers in the tissue model, physiological tissue parameters are uniformly sampled from a prior distribution. Additionally, the absorption coefficient $\mu_{a}$ and scattering coefficient $\mu_{s}$ are calculated based on the sampled values and used as input for the simulation of the light propagation. Prior distributions and $\mu_{a}$ and $\mu_{s}$ calculation are identical to \cite{C2}. The size of the tissue comprised 20 x 20 x D voxels, with a voxel size of 0.01 mm. The depth D of the tissue depends on the sampled thickness value for each layer.\par

Simulations were performed using a GPU-accelerated Python implementation of the MCX software \cite{C13}. A uniform planar light source was defined above the tissue and the number of photons at launch was set to $1 \cdot 10^{5}$. A total of $6.4 \cdot 10^5$ reflectance spectra were simulated at wavelengths 440 nm to 640 nm in steps of 4 nm, see Fig. \ref{fig:framework}b. The diffuse reflectance obtained at the surface was averaged resulting in a reflectance spectra $r(\lambda)$ of 51 wavelengths.\par

Labeling the oxygenation value from a multi-layered tissue is performed using the method proposed by Ayala et al. \cite{C14}. During every simulation, the penetration depth $p(\lambda)$ is calculated as the depth inside the tissue at which the intensity of incident light at wavelength $\lambda$ drops to a ratio of $\frac{1}{\mathrm{e}}$. A weighted average oxygenation is then calculated based on $p(\lambda)$ and the sampled thickness for every layer.\par

The obtained reflectance spectra $r(\lambda)$ from the simulations are independent of the imaging system. These are adapted to resemble measurements using a multispectral camera (Fig. \ref{fig:framework}c):

\begin{equation}
    r(b) =  \frac{w(b) + \int_{440 nm}^{640 nm} r(\lambda) \cdot L(\lambda) \cdot T(\lambda) \cdot s(b,\lambda) \,d\lambda}{\int_{440nm}^{640nm}L(\lambda) \cdot T(\lambda) \cdot s(b,\lambda) \,d\lambda}
\label{eq:band-reflectance}
\end{equation}
with $r(b)$ denoting the reflectance at band $b$ of the multispectral camera, 
$w(b)$ the camera noise for band b, $L(\lambda)$ the relative irradiance of the light source, $T(\lambda)$ the transmission profile of optical components, and $s(b,\lambda)$ the spectral response curve for band b. Finally, the spectra were normalized by dividing their area under the curve to compensate for changes in imaging distance and angles between the camera and tissue.

\subsection{Clinical data}
\label{sec:clinical-data}
Spectral data was acquired for 17 patients scheduled for elective laparoscopic esophagectomy. The surgeon visually identified and marked three regions of interest (ROIs) on the stomach graft. One corresponding to the chosen anastomotic site. The second being a well-perfused region located 2-3 $cm$ proximal from the pylorus. In addition, a small piece of tissue excised from the proximal stomach acted as an ischemic control. In 11 patients, capillary lactate was measured at every ROI using a handheld lactate analyzer (EDGE Analyzer, Apex Biotechnology, Taiwan) to further assess hypoxia.\par

Video-rate MSI was performed immediately before the lactate measurements. The XIMEA SNAPSHOT VIS (Imec, Belgium), a multispectral snapshot camera capturing images at 16 different wavelengths in the 460 - 600 $nm$ spectral range was used. The camera was mounted to a standard laparoscope using a C-mount coupler (C-mount zoom adapter, RVA Synergies, UK), and connected to a Xenon light source (CLV-S190, Olympus, Japan). During imaging, all three ROIs were within the camera's field of view. The study was approved by the ethical committee of the Ghent University Hospital (B670201836427) and informed consent was obtained from all patients. The trial is registered at Clinicaltrials.gov with registration number NCT03587532.\par

A radiometric calibration method was employed \cite{C7}. In short, the raw images underwent dark subtraction and demosaicking was performed using an algorithm developed by Muszyński and Luong \cite{C8}. Only in the case of linear unmixing, each pixel spectrum was multiplied with the correction matrix supplied by the manufacturer. Values were further divided by the spectra of the light source, obtaining reflectance data. The spectra were further normalized by dividing by their area under the curve. To obtain unlabeled real spectra later used for training the domain-adversarial models, a single frame was selected for every patient and the stomach graft was manually segmented.\par

\section{Methods}
\subsection{Deep learning models}
Four deep learning models are trained. Model architectures were kept small to ensure real-time oxygenation estimation. First, a fully connected (FCN) and convolutional neural network (CNN) were trained without a domain-adversarial approach. The FCN consists of three hidden layers containing 64, 128 and 256 nodes. In the CNN architecture, 2 convolutional layers with sizes 16 and 32 are introduced, with a kernel size of 2 performing convolutions in the spectral domain. Each layer is followed with a ReLU activation function, batch normalization and a dropout rate of 20\%. The output layer is a single linear layer estimating an oxygenation value.\par

To construct the domain adversarial networks, the same 2 model structures are used as before with the addition of a domain discriminator: a domain-adversarial fully connected network (DA-FCN) and a domain-adversarial convolutional network (DA-CNN). These models produce deep features in their hidden layers acting as a generator. A linear output layer, the discriminator, is added to the network classifying whether the produced features from the generator originate from a simulated or real spectrum.

\subsection{Experiments}
In total, $6.4 \cdot 10^{5}$ labeled spectra were simulated and adapted to the used MSI system. They were split into a training (80\%) and validation (20\%) set, and stratified based on the oxygenation. Before every epoch, the training data was augmented by simulating camera noise by means of additive Gaussian noise. The noise was added to every band b by means of the $w(b)$ parameter in Eq. \ref{eq:band-reflectance}: $w(b) \sim \eta(0, \sigma^{2})$, where $\sigma$ is adjusted to obtain an SNR of 40 dB. The validation data was similarly augmented, but only once at the start of the training. Training of the FCN and CNN was performed using the Mean Squared Error (MSE) loss. The learning rate was set to $10^{-3}$, and scheduled to reduce by a factor of 10 if the validation loss did not decrease for 10 epochs by as much as $1\%$.\par

In the case of DA-FCN and DA-CNN, unlabeled real spectra were added to train both the generator and discriminator. A total of $1.2 \cdot 10^{5}$ spectra were derived from the 6 patients without lactate measurements and split into a training set ($80\%$) and a validation set($20\%$). A total of 7328 spectra were derived from 20x20 pixel ROIs of the 11 patients with a corresponding lactate measurement, forming the separate test set. During training, a balanced sampler is used to ensure equal representations of simulated and real spectra. A domain label of 0 was given to simulated spectra and 1 to real spectra. A combined loss function was used incorporating the MSE for the regression task ($L_{r}$) and Binary Cross Entropy (BCE) loss as adversarial loss for the domain classification task ($L_{a}$ and $L_{D}$):

\begin{equation}
    L = L_{G} + L_{D} = (L_{r} + \lambda_{a}L_{a}) + L_{D}
\end{equation}

with $L_{G}$ the loss used for training the generator, $L_{D}$ the loss for training the discriminator, and $\lambda_{a}$ a factor to control the contribution of the adversarial loss set to 0.25. Before computing BCE losses, the output was transformed using a sigmoid transformation. The learning rate for the generator was set to $10^{-3}$ and that of the discriminator was to $10^{-6}$, with the same schedulers as before. Every model was trained with an Adam optimizer with a weight decay of $10^{-6}$, a batch size of 512, and for 100 epochs.

\subsection{Clinical validation}
Retrospective clinical validation is performed on the test set containing 11 patients with a corresponding capillary lactate measurement. The oxygenation quantification methods are evaluated by analyzing the relationship between oxygenation and capillary lactate, a clinical marker of hypoxia. The relationship between lactate and oxygenation is commonly described using an exponential function as follows \cite{C5}:

\begin{equation}
    Capillary \ lactate =  A \cdot e^{B \cdot O_{2}}
\label{eq:exp-fit}
\end{equation}

where $O_{2}$ corresponds to the oxygenation value and A and B are unknown fitting parameters of the exponential function.\par

The parameters A and B are determined using an ordinary least squares fit between the capillary lactate values and corresponding oxygenation estimates (averaged over a region of 20x20 pixels around the lactate sampling point). The mean absolute error (MAE) and the $R^{2}$ value of the fit, and the correlation coefficient, are selected to assess the performance of the different oxygenation quantification methods. The higher the correlation with lactate and the better the fit with the exponential function, the better the relation between predicted oxygenation and the hypoxia marker capillary lactate. In addition, these values are compared to those obtained by predicting oxygenation using a linear mixing approach outlined in previous work \cite{C9}.\par  

\section{Results}
\begin{figure}
    \centering
    \begin{subfigure}{0.48\textwidth}
        \centering
        \includegraphics[width=\textwidth]{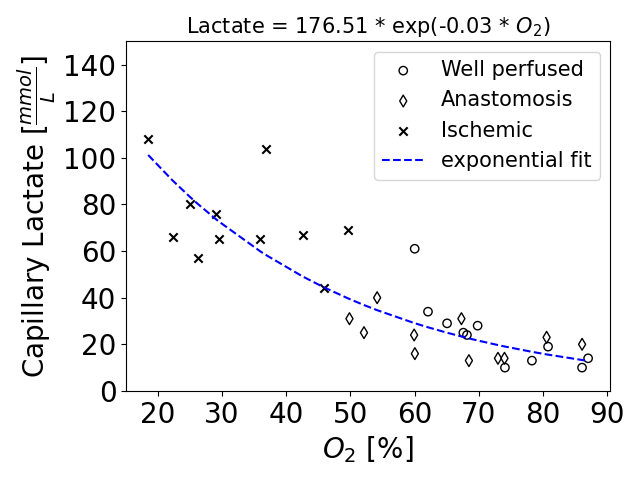}
        \caption{DA-FCN}
    \end{subfigure}
    \hspace{0.2cm}  
    \begin{subfigure}{0.48\textwidth}
        \centering
        \includegraphics[width=\textwidth]{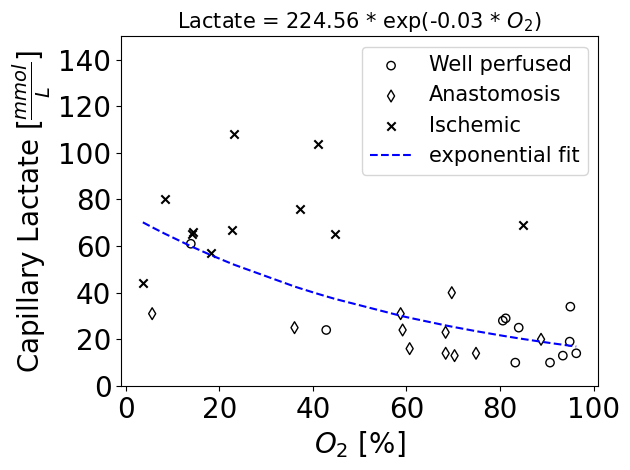}
        \caption{Linear unmixing}
    \end{subfigure}
    \caption{The exponential fit between the capillary lactate measurements and the predicted oxygenation of the best model (a) and linear unmixing (b).}
    \label{fig:correlation}
\end{figure}

\begin{figure}[H]
    \centering
    \begin{subfigure}{0.3\textwidth}
        \centering
        \includegraphics[width=\textwidth]{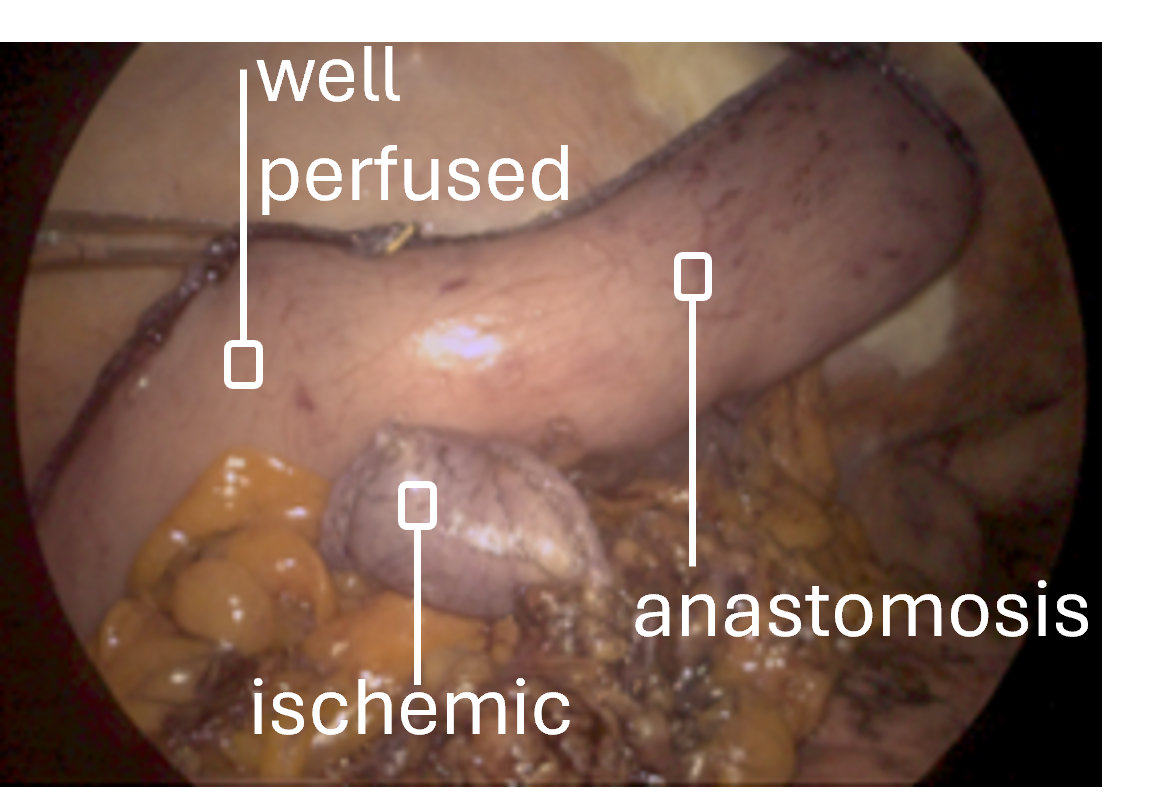}
        \caption{Synthetic RGB}
    \end{subfigure}%
    \hspace{0.05cm}  
    \begin{subfigure}{0.3\textwidth}
        \centering
        \includegraphics[width=\textwidth]{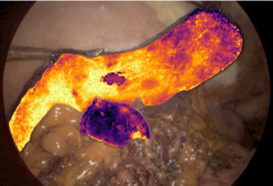}
        \caption{Linear unmixing}
    \end{subfigure}%
    \hspace{0.05cm}  
    \begin{subfigure}{0.3\textwidth}
        \centering
        \includegraphics[width=\textwidth]{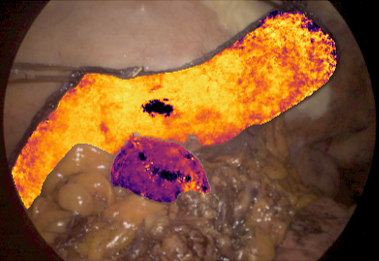}
        \caption{FCN}
    \end{subfigure}

    \vspace{0.1cm}  
    \begin{subfigure}{0.3\textwidth}
        \centering
        \includegraphics[width=\textwidth]{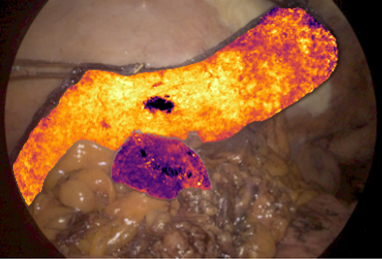}
        \caption{CNN}
    \end{subfigure}%
    \hspace{0.05cm}  
    \begin{subfigure}{0.3\textwidth}
        \centering
        \includegraphics[width=\textwidth]{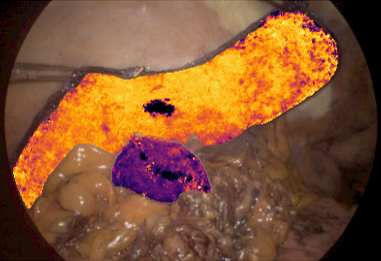}
        \caption{DA-FCN}
    \end{subfigure}%
    \hspace{0.05cm}  
    \begin{subfigure}{0.3\textwidth}
        \centering
        \includegraphics[width=\textwidth]{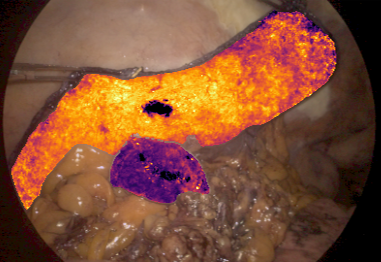}
        \caption{DA-CNN}
    \end{subfigure}
    \caption{(a) shows a synthetic RGB image of a patient included in the clinical test set, along with the selected ROIs of the well-perfused, anastomosis and ischemic tissue, (b) Oxygenation maps produced by linear unmixing and machine learning methods. The scale ranges from 0\% (dark purple) to 100\% (bright yellow).}
    \label{fig:oxygenation_maps}
\end{figure}

The performance of different models for oxygenation quantification is summarized in Table \ref{tab:metrics}, containing the obtained training and validation regression losses, along with the evaluation metrics from the clinical validation. In general, the deep learning approaches show an improved correlation with lactate measurements compared to the linear unmixing technique. Moreover, the addition of adversarial training to the deep learning models further improves correlations with lactate resulting in clinically improved oxygenation estimations. The best performing model on the clinical test set is the DA-FCN. Fig. \ref{fig:correlation} shows the exponential fit of the capillary lactate with the oxygenation predictions of this model compared to linear unmixing. Fig. \ref{fig:oxygenation_maps} shows oxygenation maps of the stomach graft from a single multispectral image from the clinical test data. All models successfully estimate a high oxygenation for the well-perfused tissue and critically low oxygenation for the ischemic tissue. Furthermore, the anastomosis site seems to lie at a borderline, as a transition into lower oxygenated tissue can be seen. The deep learning models estimate less extreme, and more realistic, values compared to linear unmixing.

\begin{table}[ht]
\setlength{\tabcolsep}{6pt} 
\renewcommand{\arraystretch}{1.3} 
\centering
\resizebox{\textwidth}{!}{%
\begin{tabular}{|c|cc|cccc|}
\hline
 & \begin{tabular}[c]{@{}c@{}}Train \\ loss ($\cdot 10^{-3}$)\end{tabular} & \begin{tabular}[c]{@{}c@{}}Validation \\ loss ($\cdot 10^{-3}$)\end{tabular} & \begin{tabular}[c]{@{}c@{}}MAE  \\ {[}mmol/L{]}\end{tabular} & \begin{tabular}[c] {@{}c@{}}$R^{2}$  \\ {[}a.u.{]}\end{tabular} & \begin{tabular}[c] {@{}c@{}}Correlation  \\ {[}a.u.{]}\end{tabular} & \begin{tabular}[c]{@{}c@{}}Inference time \\ {[}ms{]}\end{tabular}\\ \hline
FCN & \textbf{6.47} & \textbf{9.59} & $10.81 \pm 10.73$ & 0.74 & -0.86 & \textbf{16.64}  \\
CNN & 8.80 & 9.86  & $13.33 \pm 11.98$ & 0.71  & -0.84  & 25.28 \\
DA-FCN & 8.04 & 11.78 & \textbf{9.83 $\pm$ 10.05} & \textbf{0.76} & \textbf{-0.87}  & 16.80 \\
DA-CNN & 11.04 & 12.06 & $12.26 \pm 10.54$ & 0.74  & -0.86 & 25.88 \\ \hline
Linear unmixing &  &  & $15.32 \pm 15.94$ & 0.46 & -0.68 & 94.68 \\ \hline
\end{tabular}
}
\caption{Model training and validation regression losses, and their performance on the clinical test data with corresponding lactate measurements. In addition, the average inference times over 1000 iterations on a multispectral image are shown using an NVIDIA GeForce RTX 4060 Laptop GPU.}
\label{tab:metrics}
\end{table}

\section{Discussion}
Deep learning models improve oxygenation estimation compared to traditional linear unmixing, showing a stronger correlation with the hypoxia marker capillary lactate. Despite higher training loss, domain-adversarial training further enhances performance by reducing the domain gap between simulated and real data. This occurs because the generator learns deep features optimized for oxygenation estimation while maintaining domain invariance through adversarial training. As a result, these models demonstrate greater robustness, leading to improved performance on the clinical test set. \par

Both deep learning and linear unmixing detect large oxygenation differences, with linear unmixing showing more extreme values, particularly between well-perfused and ischemic tissue. However, their estimates diverge in regions with subtle differences, such as the anastomosis. As shown in Fig. \ref{fig:correlation}b, linear unmixing tends to underestimate oxygenation in the anastomosis and overestimate it in ischemic tissue, leading to outliers in the fit. Deep learning mitigates these errors, resulting in a more accurate correlation. \par

Additionally, deep learning models achieve oxygenation mapping at 50–62 frames per second on a low-end laptop GPU, which is 5–6 times faster than linear unmixing. This makes them a promising tool for intraoperative guidance based on real-time oxygenation estimation. \par

As the study is limited by the number of patients and tissue types included, future work will enhance statistical significance and robustness by addressing this. Additionally, we are exploring the transformation of 1D simulated spectra into 2D data to incorporate spatial information.

\section{Conclusion}
Deep learning methods, particularly domain-adversarial networks, improve oxygenation estimation compared to traditional linear unmixing. Our results demonstrate the potential of deep learning to translate spectral imaging into a clinically viable tool for real-time oxygenation assessment, showing strong correlations with a hypoxia marker. 

\section{Acknowledgments}
This work has received funding from the Flemish Government under the “Onderzoeksprogramma Artificiele Intelligentie (AI) Vlaanderen” programme.
\newpage

\bibliographystyle{splncs04}
\bibliography{main}

\end{document}